\documentclass[11pt,]{article}
\usepackage{charter}
\usepackage{amssymb,amsmath,graphicx,url}
\usepackage{setspace,upgreek,amsmath}
\usepackage{geometry}[margin=0.75in]
\usepackage{lineno}
\usepackage{placeins}
\usepackage[sort&compress,numbers,comma,square]{natbib}

\title{Quantifying sources of uncertainty in drug discovery predictions with probabilistic models}
\author{Stanley E. Lazic\textsuperscript{1,*}, Dominic P.
Williams\textsuperscript{2}}
\date{}

\begin{document}
\maketitle

\textsuperscript{1}\emph{Prioris.ai Inc., 459--207 Bank Street, Ottawa,
K2P 2N2, Canada}

\textsuperscript{2}\emph{Functional and Mechanistic Safety, Clinical
Pharmacology and Safety Sciences, AstraZeneca, R\&D, Cambridge, CB4 0WG,
UK}

\textsuperscript{*}Corresponding author: \texttt{stan.lazic@cantab.net}

\section*{Abstract}
Knowing the uncertainty in a prediction is critical when making expensive investment decisions and when patient safety is paramount, but machine learning (ML) models in drug discovery typically provide only a single best estimate and ignore all sources of uncertainty. Predictions from these models may therefore be over-confident, which can put patients at risk and waste resources when compounds that are destined to fail are further developed. Probabilistic predictive models (PPMs) can incorporate uncertainty in both the data and model, and return a distribution of predicted values that represents the uncertainty in the prediction. PPMs not only let users know when predictions are uncertain, but the intuitive output from these models makes communicating risk easier and decision making better. Many popular machine learning methods have a PPM or Bayesian analogue, making PPMs easy to fit into current workflows. We use toxicity prediction as a running example, but the same principles apply for all prediction models used in drug discovery. The consequences of ignoring uncertainty and how PPMs account for uncertainty are also described. We aim to make the discussion accessible to a broad non-mathematical audience. Equations are provided to make ideas concrete for mathematical readers (but can be skipped without loss of understanding) and code is available for computational researchers (\url{https://github.com/stanlazic/ML_uncertainty_quantification}).

\clearpage

\section*{Introduction}

At each stage of the drug discovery pipeline, researchers decide to progress or halt compounds using both qualitative judgements and quantitative methods. This is formally a prediction problem, where, given some information, a prediction is made about a future observable outcome. Standard predictive or machine learning models such as random forests, support vector machines, or neural networks only report point-estimates, or a single ``best'' value for a prediction; they provide no information on the uncertainty of the prediction. Prediction uncertainty is important when (1) the range of plausible values is as important as the best estimate, (2) you need to know that the model cannot confidently make a prediction, (3) you need to reliability distinguish between ranked items, or (4) the cost of an incorrect decision is large; for example, when making expensive investment decisions or when assessing patient safety.

\begin{figure}[ht]
\centering
\includegraphics[scale=0.6]{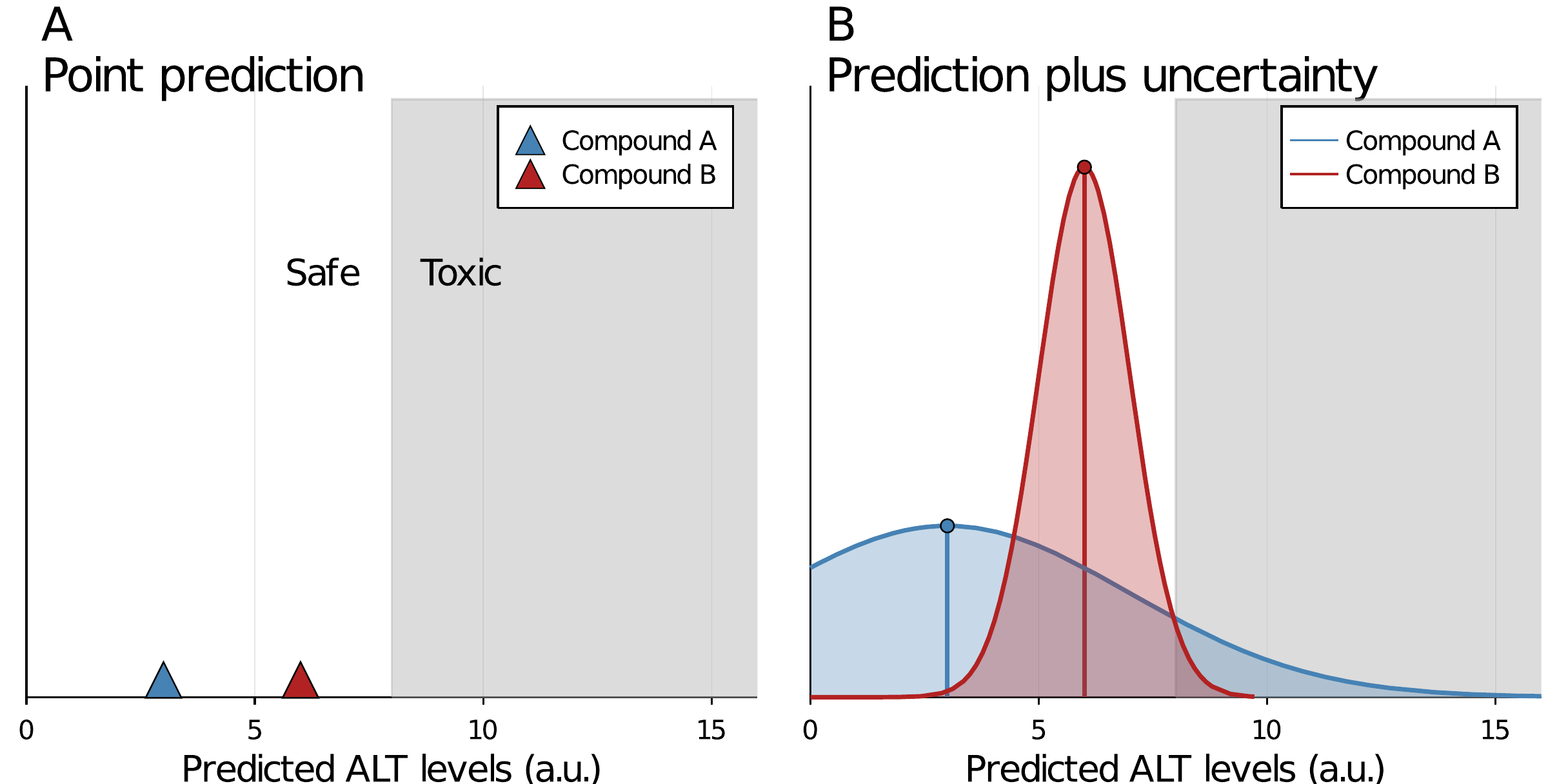}
\caption{Prediction uncertainty. Predicted blood ALT levels for two compounds, with Compound A appearing better (A). Values above the safety threshold of 8 (grey shaded region) will not be progressed. Reporting prediction uncertainty shows that 18\% of Compound A is above the threshold but only 2\% of Compound B, indicating that Compound B is better (B).  ALT = alanine aminotransferase; a.u. = arbitrary units.}
  \label{fig:drug_example}
\end{figure}

Figure \ref{fig:drug_example} shows predicted clinical blood alanine aminotransferase (ALT) levels -- an indicator of liver toxicity --  for two hypothetical compounds. Assume that levels below 8 (arbitrary units) are considered safe, and that only one compound can be taken forward for clinical trials. Based only on the best estimate, compound A (blue) is preferable (Fig. \ref{fig:drug_example}A). Knowing the prediction uncertainty changes the picture (Fig. \ref{fig:drug_example}B). 14\% of compound A's distribution is in the unsafe shaded region, while only 2\% of compound B's distribution is in the shaded region. Based on these distributions, compound B maybe the better candidate to progress to clinical trials; or a project team may decide to run more experiments to reduce compound A's uncertainty.

We define a probabilistic predictive model (PPM) as any machine learning model that returns a distribution for the prediction instead of a single value. PPMs differ in how they represent uncertainty. At one end, fully probabilistic Bayesian models specify a distribution for the outcome and for all unknown parameters in the model. These models are the gold-standard for quantifying uncertainty but they can be computationally expensive. At the other end are method that return multiple predicted values without specifying a probability distribution. Examples include fitting the same model on multiple bootstrapped datasets, or for models with a stochastic component, fitting the same model using different random number generator seeds \cite{Lakshminarayanan2017}. Although these models are often computationally tractable, the connection between the distribution of predicted values and uncertainty is unclear. For example, does varying something trivial such as the random number generator seed adequately capture our uncertainty in a prediction? Between these extremes are approaches that try to obtain the benefits of fully probabilistic models using approximations, reformulations, or computational shortcuts. For example, instead of using a fully Bayesian deep neural network, Kristiadi et al. were able to obtain many of the benefits by making only the final layer of the network Bayesian \cite{Kristiadi2020}. Also in the neural network literature, Gal and Ghahramani approximated parameter uncertainty by randomly inactivating nodes at prediction time -- a procedure known as Monte Carlo dropout \cite{Gal2016}, and Teye and colleagues used the mean and variances calculated from batch normalisation steps for the same purpose  \cite{Teye2018}. These approximate methods are an active area of research \cite{Khosravi2011,Welling2011,Blundell2015,Pearce2020} (see Mervin et al., for a recent review \cite{Mervin2021}), and will be critical for making PPMs more widely adopted, but here we focus on fully probabilistic models as they better highlight the key areas of uncertainty. 

All prediction models can be written as

\begin{equation}
  y\, |\, x
\end{equation}

\noindent and read as ``$y$ given $x$'' -- where $y$ is the outcome to be predicted and $x$ are one or more variables used to predict $y$. Both $x$ and $y$ are available when training a model, and when a model is deployed, new $x$ values are observed and used to predict the unknown $y$'s. PPMs additionally provide a probability distribution for $y$, denoted as

\begin{equation}
  \label{eq:cond_prob}
  P(y\, |\, x)
\end{equation}

\noindent and read as ``the probability of $y$ given $x$''. A distribution for $y$ enables us to calculate any metric of interest, such as the best guess for $y$ (e.g. mean, median, or mode), thereby providing the same information as standard methods. But in addition, prediction intervals (PI) can be calculated around the best estimate, or the probability that $y$ is greater or less than a predefined threshold can be calculated, as was done in Figure \ref{fig:drug_example}B.

Below we describe the sources of uncertainty and the advantages of PPMs. We aim to make the discussion accessible to a broad non-mathematical audience. Equations are provided to make ideas concrete for mathematical readers, but can be skipped without loss of understanding of the remaining text. In addition, code is provided for computational researchers (\url{https://github.com/stanlazic/ML_uncertainty_quantification}) and implemented in Julia using Turing \cite{Bezanson2017,Ge2018}.

\section*{Sources of uncertainty}

The definition of a probabilistic model in Equation \ref{eq:cond_prob} lacked a crucial component, which is the model itself:

\begin{equation}
P(y\, |\, x, \mathrm{Model}).
\end{equation}

The sources of uncertainty are now clear: they can reside in the data ($x$, $y$) or in the model. A prediction for $y$ not only depends on $x$, but on the model used to connect $x$ and $y$. Hence, predictions are conditional on a model, and uncertainty in the model should lead to greater uncertainty in the prediction. Models are composed of the following components, which we discuss in greater detail below:

\begin{enumerate}
\item \textbf{Data.} The outcome $y$ and the predictor or input variables $x$.

\item \textbf{Distribution function.} The distribution that represents our uncertainty in $y$, also called the likelihood or data generating distribution: $G(\cdot)$.
  
\item \textbf{Mean function.} The functional or structural form of the model describing how $y$ changes as $x$ changes: $f_{\mu}(\cdot)$.
  
\item \textbf{Variance function.} Describes how the uncertainty in $y$ varies with $x$: $f_{\sigma}(\cdot)$.
  
\item \textbf{Parameters.} The unknown coefficients or weights for the mean ($\theta_{\mu}$) and variance ($\theta_{\sigma}$) functions that are estimated from the data.

\item \textbf{Hyperparameters.} Parameters or other options used to define a model that are not estimated from the data but fixed or selected by the analyst: $\phi$.
  
\item \textbf{Link functions.} Nonlinear transformations of the mean ($l_{\mu}(\cdot)$) and/or variance function ($l_{\sigma}(\cdot)$) used to keep values within an allowable range.
\end{enumerate}

Combining these seven components gives the generic formulation of a PPM (Eq. \ref{eq:PPM}). Although this equation is abstract, it captures where uncertainty can reside. In this section we carefully describe the terms in the equation and then provide a concrete example.

\begin{align}
  \label{eq:PPM}
  y & \sim G(\mu, \sigma) \\  \nonumber
  \mu & = l_{\mu}(f_{\mu}(x; \theta_{\mu})) \\  \nonumber
  \sigma & = l_{\sigma}(f_{\sigma}(x; \theta_{\sigma}))
\end{align}

Starting with the first line of Equation \ref{eq:PPM}, $y$ is a future value that we want to predict and it could represent a clinical outcome, an IC$_{50}$ value, or a physicochemical property of a compound such as solubility. PPMs require that we specify a distribution for our uncertainty in $y$, which we can informally think of as the distribution from which $y$ was generated. We denote this distribution as $G(\cdot)$ and the ``$(\cdot)$'' notation indicates that $G$ is a function with inputs ($\mu$ and $\sigma$ in this case), but places the focus on $G$ and not the inputs, thereby reducing clutter. Common distributions include the normal/Gaussian (for continuous symmetric data), Student-t (continuous symmetric data with outliers), Bernoulli (0/1 data), and Poisson (count data). The mean of the chosen distribution is given by $\mu$ and many distributions have a second parameter, $\sigma$, that controls the spread or width of the distribution. This parameter is critical for PPMs because it describes the uncertainty in $y$. The $\sim$ symbol can be read as ``is distributed as'' or ``is generated from''. To make this concrete, we might represent our uncertainty in $y$ for a given compound with a Gaussian distribution that has a mean $\mu = 2.45$ -- which represents our best estimate of $y$ -- and a standard deviation of $\sigma = 2.1$. This would be written as $y \sim \mathrm{Normal}(2.45, 2.1)$.

But where did our best estimate $\mu$ come from? This is defined on the second line of Equation \ref{eq:PPM}. $x$ is the input data used to predict $y$ and it could represent the compound structure (encoded as a binary fingerprint for example), assay results, or physicochemical properties. The prediction task reduces to using $x$ to predict $y$, but they need to be connected through a statistical or machine learning model. The structural or functional form of this model is denoted by $f_{\mu}(\cdot)$ and it could represent the structure of a simple linear regression model, the architecture of a neural network, an ensemble of trees, or a differential equation representing a pharmacokinetic model. We refer to $f_{\mu}(\cdot)$ as the \textit{mean function} because it tells us the predicted mean value of $y$ for a given value of $x$. The mean function contains parameters ($\theta_{\mu}$) that are estimated from the data, and the parameters could represent the coefficients of a linear model or the weights and biases of a neural network. The ``learning'' in machine learning refers to estimating values for the parameters that maximises predictive performance, given the data and functional form of the model. $\theta_{\mu}$ usually represents multiple parameters in the mean function; for example, it would represent both the intercept and slope in a simple regression model. The subscript $\mu$ on $f$ and $\theta$ indicates that the function and parameters refer to the mean, since we also have a function and parameters for the variance, $\sigma$, which we described further below.

One problem is that $\mu$ has no constraints, and can be any value calculated from  $f_{\mu}(\cdot)$, which may lead to impossible predictions. For example, if we're predicting the probability that a compound is toxic, $f_{\mu}(\cdot)$ needs to be between zero and one -- values outside this range do not make sense. Hence, we need to transform $f_{\mu}(\cdot)$ with a \textit{link function} $l_{\mu}(\cdot)$ to put it within a permissible range. One option is to use the equation $1/(1 + \mathrm{exp}(-f_{\mu}(\cdot)))$ as a link function, which compresses $f_{\mu}(\cdot)$ into the 0-1 range, but other functions are also possible. A link function is unnecessary when no restriction on $f_{\mu}(\cdot)$ is required, and $l_{\mu}(\cdot)$ can be dropped from the formula. (Useful mnemonics: $f$ = \underline{F}unction; $G$ = data \underline{G}enerating distribution; $l$ = \underline{L}ink function; with subscripts $\mu$ and $\sigma$ referring to the mean and variance, respectively).

The third line of Equation \ref{eq:PPM} shows that the variance or uncertainty in a predicted value of $y$ can be specified in the same way as we specify the mean or best estimate of $y$. Many models assume a constant value for $\sigma$, which is the ``homogeneity of variance'' assumption in traditional statistical models. However, a model's uncertainty in $y$ may vary for different values of $x$, which can be captured with a \textit{variance function} $f_{\sigma}(\cdot)$. Since variances are positive, a link function for the variance ($l_{\sigma}(\cdot)$) is needed to constrain $\sigma$ to be greater than zero.

Fully Bayesian PPMs also require us to specify the uncertainty in the parameters $\theta_{\mu}$ and $\theta_{\sigma}$ before analysing the data, and the hyperparameters ($\phi$) refer to this specification. Many of the approximate methods avoid this step and $\phi$ may not correspond to any part of the model but defines other options for the learning algorithm and therefore it is not included in Equation \ref{eq:PPM}.

\begin{figure}[ht]
\centering
\includegraphics[scale=0.6]{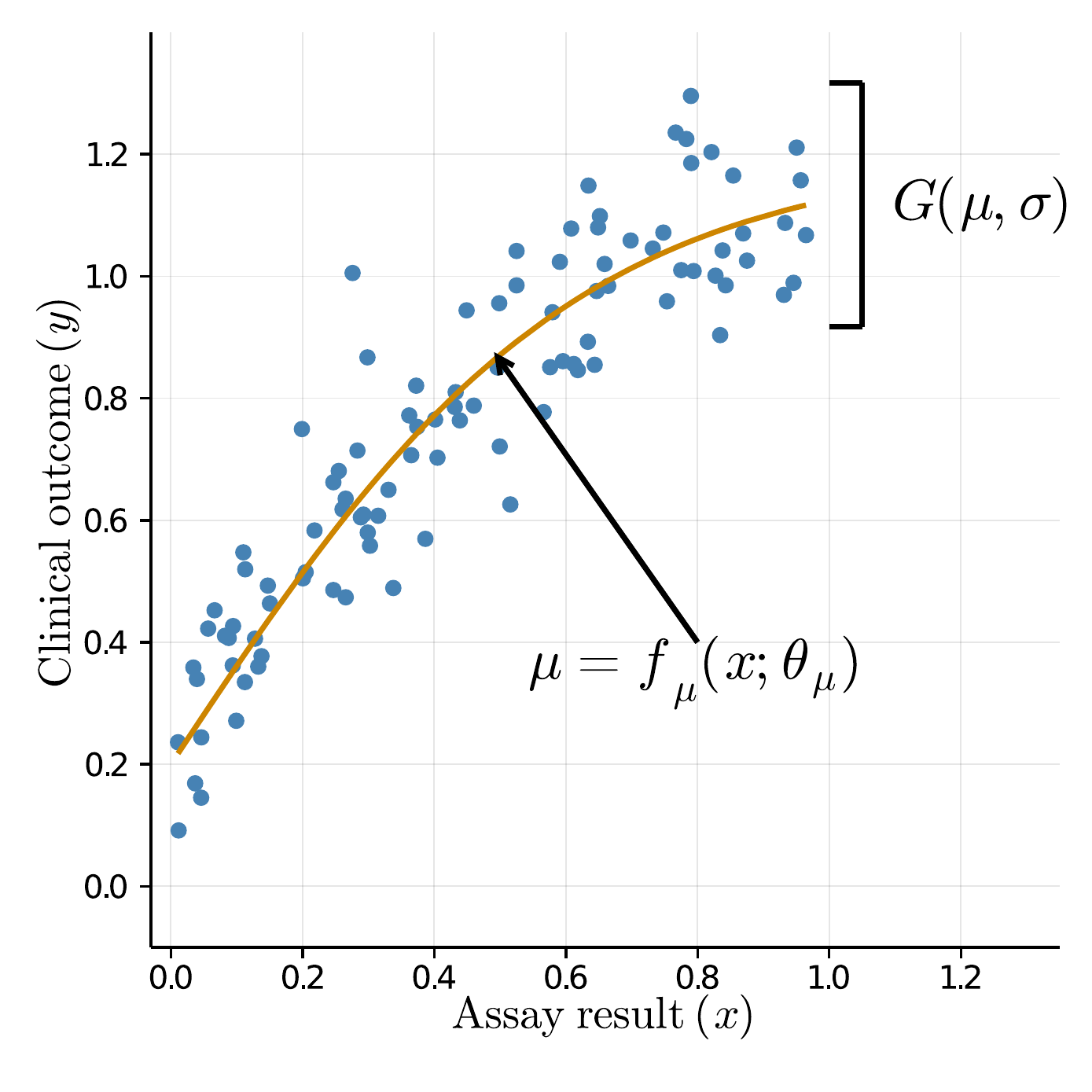}
\caption{Simulated data with the true relationship between $x$ and $y$ given by $\mu = f_{\mu}(x; \theta_{\mu})$ (orange curve) and based on Equation \ref{eq:true}. $G$ is the Gaussian data generating distribution with a mean $\mu$ and a constant variance $\sigma$, which models the spread of points around the line. Link functions for $\mu$ and $\sigma$ are not used and hence are not shown.}
  \label{fig:sim_dat}
\end{figure}

Usually the $x$ and $y$ data are all we have, and we need to choose $G$, $l_{\mu}$, $f_{\mu}$, $l_{\sigma}$, $f_{\sigma}$, and $\phi$ based on background knowledge, preliminary plots of the data, or trying several options and empirically assessing which is best. For example, an initial model might assume that $G$ is Gaussian and that $f_{\mu}$ follows a hypothesised mechanistic relationship based on a pharmacokinetic model. Then, we estimate or learn the values of $\theta_{\mu}$ based on training data, and assess the prediction on a separate test data set. To make these ideas concrete, Figure \ref{fig:sim_dat} shows simulated data for 100 compounds, where $y$ is a clinical outcome and $x$ is an assay result. Assume that higher values of $y$ indicate greater toxicity.

A nice feature of PPMs is that they are generative, meaning that they can generate or simulate data. Indeed, simulation and learning are opposite sides of the same coin: learning takes the fixed data and infers likely values of the parameters that could have generated the data, whereas simulation fixes the parameters and generates the data. The model in Equation \ref{eq:true} generated the data in Figure \ref{fig:sim_dat} and we will use it as a running example throughout

 \begin{align}
   \label{eq:true}
   y & \sim \mathrm{Normal}(\mu , \sigma) \\
   \mu & = \theta_2 + \dfrac{1 - e^{-\theta_1 x}}{1 + e^{-\theta_1 x}}. \nonumber
 \end{align}

The first line of the equation is read as: ``the outcome $y$ is generated ($\sim$) from a Gaussian or Normal distribution with a mean of $\mu$ and a standard deviation of $\sigma$''. But how does $y$ depend on $x$? The second line shows how $x$ enters and how it depends on two parameters: $\theta_1$ and $\theta_2$ ($e$ is a constant, not a parameter). We set this equation equal to $\mu$ and can substitute it for $\mu$ in the first line of Equation \ref{eq:true} giving

 \begin{equation*}
   y  \sim \mathrm{Normal}\left(\theta_2 + \dfrac{1 - e^{-\theta_1 x}}{1 + e^{-\theta_1 x}}\ , \sigma \right).
 \end{equation*}

Writing the equation in one line makes the relationship between $x$ and $y$ clearer, but multi-lined equations are easier to read with more complex models. For most prediction models the parameters are uninteresting, but to help interpret the model, $\theta_2$ is the $y$-intercept (value of $y$ when $x = 0$), and $\theta_1$ controls how quickly the line in Figure \ref{fig:sim_dat} reaches the upper limit of $y = 1.2$ as $x$ gets large.

To simulate a value for $y$, we need to (1) select parameter values, and we use the following: $\theta_1 = 3.25$, $\theta_2 = 0.2$, and $\sigma = 0.1$; (2) select a value of $x$, which enables us to calculate $\mu$; then (3) draw a random number from a Gaussian distribution with a mean of $\mu$ and standard deviation of $\sigma$. This can be repeated any number of times to obtain the $P(y|x)$ distribution, and for different values of $x$. The data in Figure \ref{fig:sim_dat} were generated for 100 $x$ values uniformly distributed between 0 and 1. Note that $\theta_{\mu}$ in Equation \ref{eq:PPM} is a place-holder for several variables, which correspond to $\theta_1$ and $\theta_2$ in Equation \ref{eq:true}. Given this data, we now illustrate were where the seven sources of uncertainty enter. 

\subsection*{Mean function uncertainty}

Uncertainty in the mean function $f_{\mu}(\cdot)$ arises because we rarely know the true form of the relationship between $x$ (assay) and $y$ (outcome) -- that is, we don't know the form of Equation \ref{eq:true}. Uncertainty in the mean function is also called model uncertainty, but this term is ambiguous because models have multiple components. Choices for the mean function include which predictors, interaction terms, transformations, basis expansions, hierarchies, and time-varying components to include in the model. Assume we only observe the data in Figure \ref{fig:sim_dat}, several models we might consider for the relationship between $x$ and $y$ are:

\begin{align*}
\mathrm{Linear:} \qquad & y  = \theta_0 + \theta_1 x  \\
\mathrm{Quadratic:} \qquad & y  = \theta_0 + \theta_1 x + \theta_2 x^2  \\
\mathrm{2\ Parameter\ Exponential:} \qquad & y =  \theta_2\ (1 - e^{-\theta_1 x})  \\
\mathrm{3\ Parameter\ Exponential:} \qquad & y =  \theta_3 + \theta_2\ (1 - e^{-\theta_1 x})\\
\mathrm{Michaelis-Menten:} \qquad & y  = \theta_1 x\ /\ (\theta_2 + x).
\end{align*}

\begin{figure}[ht]
  \centering
  \includegraphics[scale=0.4]{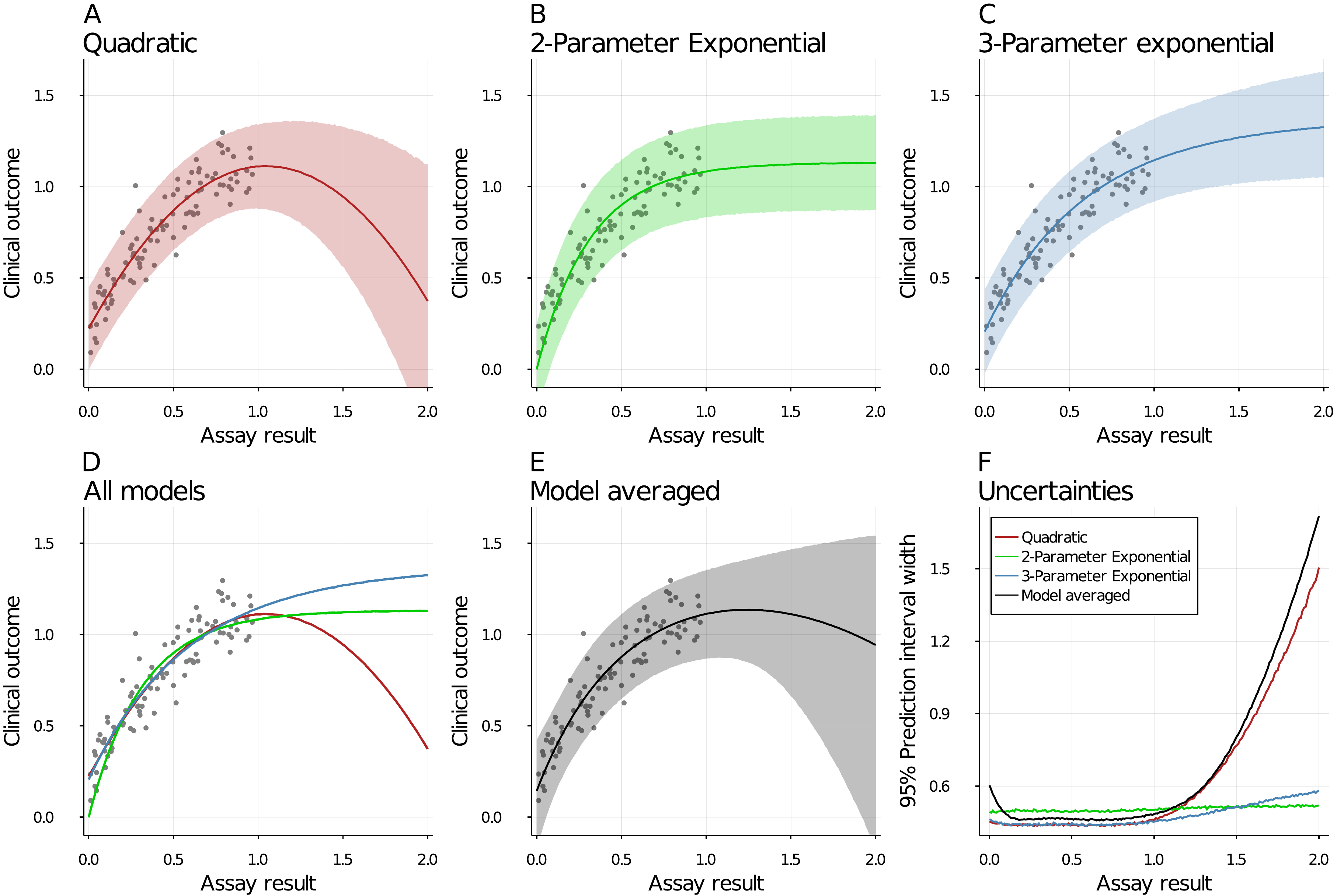}
  \caption{Model averaging. Three models fit the data well (A-C), even though none are the true model. They make different predictions at low assay values and when extrapolating to higher values. The mean predictions are superimposed for easier comparison (D). Model averaged prediction (E). Comparison of prediction interval widths (F). Shaded regions are the 95\% prediction intervals.}
  \label{fig:mod_comp}
\end{figure}

None of these are the true model (Eq. \ref{eq:true}), but except for the linear model, they all saturate at high values of $x$ or are concave and therefore capture the main trend in the data. Which model should we use? Typically, only a single model is selected and predictions are made from that. If one model is clearly better than the others, there may be little lost by using one model for predictions. However, if two or more models fit the data equally well, making predictions from only one will underestimate the prediction uncertainty. Fortunately, we are not forced to choose one model but can fit several and combine their predictions. To illustrate, we will use the quadratic, 2-parameter exponential, and 3-parameter exponential models. The three models are fit to the data (Fig. \ref{fig:mod_comp}A--C) and predictions are extrapolated to show both how similar the fits are where there is data, and how different the fits are when extrapolating. The shaded regions show the 95\% prediction intervals (PI), and if a model is suitable, we expect 95\% of the data to fall in the shaded region.

To better compare the predictions, the mean functions $\mu$ for three models are plotted together in Figure \ref{fig:mod_comp}D. The models make similar predictions within the range of the data, except at very low values of $x$, where the 2-parameter exponential model predicts smaller values. Figure \ref{fig:mod_comp}E shows the model-averaged prediction, and note how uncertainty is greater at high values of $x$, where the models make different predictions. To better appreciate how model averaging incorporates uncertainty, Figure \ref{fig:mod_comp}F shows the width of the 95\% PI for the three models and the averaged model. Note how the averaged model has a wider PI at low values of $x$ than any of the original models. This occurs because the three models make different predictions at low values and hence this region is more uncertain. For intermediate values of $x$, all three models make similar predictions and the averaged PI is wider than some models and lower than others. When extrapolating to larger values $x$, the model averaged PI width quickly becomes the widest, reflecting both the diverging predictions of the individual models and the greater uncertainty in the quadratic model.

Predictions are always conditional on a model, and if we don't know the true model, predictions from the wrong model are likely to be overconfident. We extrapolated well beyond the data to illustrate how model averaging accounts for model uncertainty. Such extrapolation may seem unrealistic, but the message is that whenever models make different predictions for the same values of $x$, the uncertainty in the model-specific predictions will be overconfident, as we see to a lesser degree for low values of the assay.

\subsection*{Parameter uncertainty}
Most predictive models have parameters or weights that are learned from the data and control how the predictor variables ($x$) are related to the outcome variable ($y$) -- these parameters are the $\theta$ symbols in the five models considered above. Most machine learning methods only use the single best value of each parameter when making a prediction. But since the parameters are learned from the data, they are uncertain, and this uncertainty should be propagated into the prediction. Parameter uncertainty decreases as the sample sizes increases, so to better illustrate the effect of parameter uncertainty on predictions, a smaller dataset was made by taking every eighth data point from the previous example.

\begin{figure}[ht]
  \centering
  \includegraphics[scale=0.5]{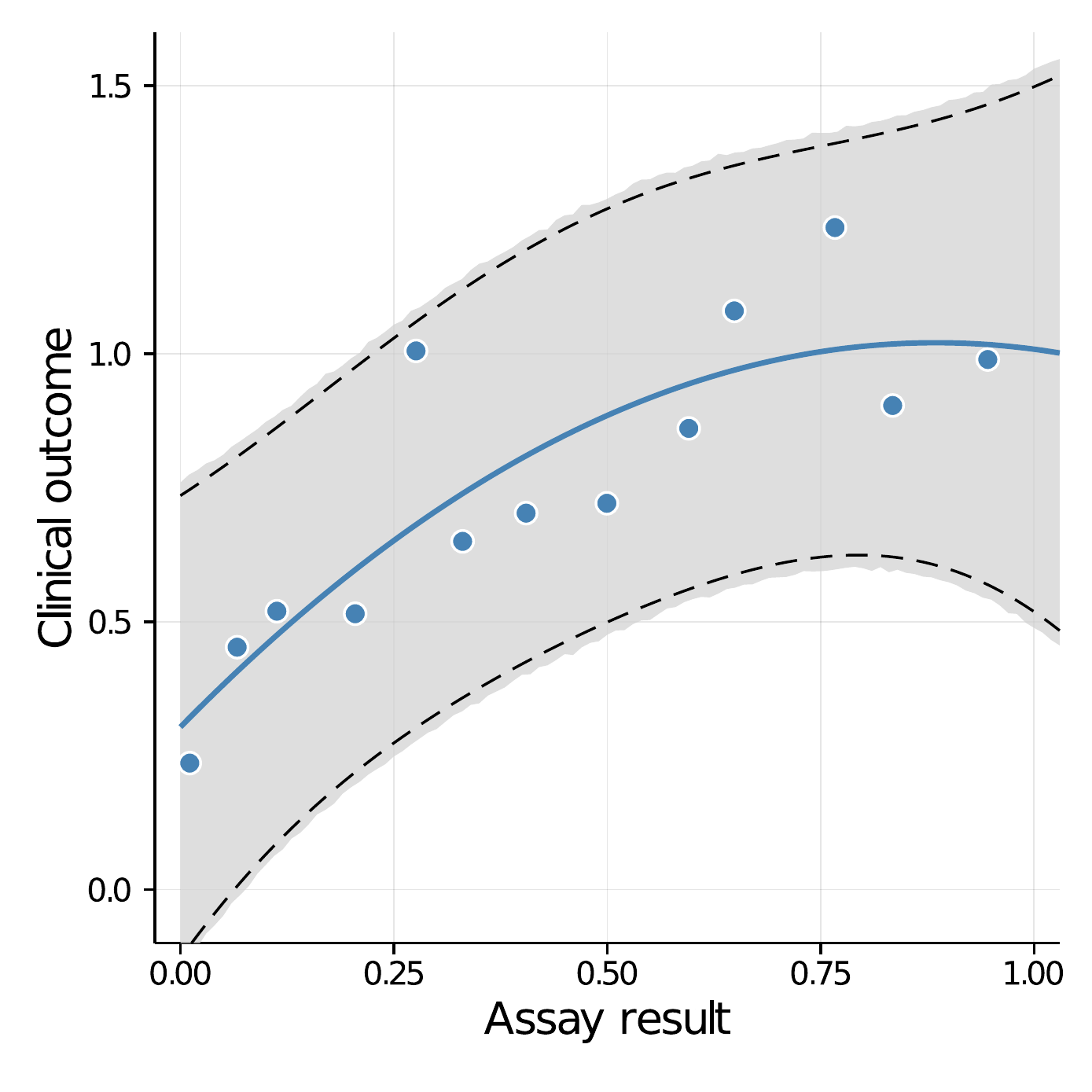}
  \caption{Parameter uncertainty. The grey shaded region is the 95\% prediction interval for a Bayesian model and the narrower black dashed lines for a classical model. The classic interval is narrower because it doesn't account for uncertainty in the parameters.}
  \label{fig:para_uncert}
\end{figure}

Figure \ref{fig:para_uncert} uses the quadratic model, which has four parameters ($\theta_0, \theta_1, \theta_2, \sigma$). The grey shaded region shows the 95\% prediction interval from a Bayesian model that accounts for parameter uncertainty. The dashed black lines show the 95\% PI from a classic quadratic regression model which ignores parameter uncertainty, and note how they are slightly narrower. The mean function is identical for both the Bayesian and classic model.

The difference in PI width may seem negligible when focusing on the mean prediction, but ignoring parameter uncertainty gives approximately 7\% narrower PIs. This may be important with  ``point of departure'' calculations when the tails of the distributions are more important than the means \cite{Reynolds2020}. Assume clinical outcomes above 1.2 are considered problematic and the assay result for one compound is $x=0.5$. When fully accounting for parameter uncertainty, there is a 5.9\% chance that the true value of the clinical outcome is above 1.2, versus a 4.3\% chance when ignoring uncertainty. The ratio of these numbers is 1.37, indicating that the tail area is nearly 1.4 times larger when accounting for uncertainty.

Models typically have many more parameters than this example (the state-of-the-art Generative Pre-trained Transformer 3 (GPT-3) deep learning language model has 175 billion parameters \cite{Brown2020}) and simply collecting more data is often not an option to reduce parameter uncertainty because more data enables more complex models to be fit (e.g. including nonlinear terms and interactions), which then increases the number of parameters.

\subsection*{Hyperparameter uncertainty}

Hyperparameters are a diverse set of tunable options that affect the training and predictions. Unlike parameters, they are not estimated from the data but selected by the analyst; examples include the amount of regularisation in a lasso model, the number of trees in a random forest model, or the cost function in a support vector machine. Suitable values are typically found by trying several options and selecting the best using crossvalidation. In the hyperparameter category we can also include options that are rarely part of a formal selection process such as the choice of optimisation algorithm or random number seed for models with a stochastic component. For fully Bayesian models we can also include parameters for prior distributions, which are not updated by the data. These hyperparameters are selected pragmatically to provide good predictions, but other sets of hyperparameter values might give equally good predictions, on average, but slightly different predictions for each test compound. Hence, uncertainty in hyperparameter values is rarely taken into account. A further complication is that most hyperparameters are not related to any biological or chemical quantity of interest and hence it is unclear what the uncertainty is actually about.  Nevertheless, Lakshminarayanan and colleagues showed that by running many models with a different random seed, the ensemble of predictions performed better than a single model, and the distribution of predicted values provided a measure of uncertainty \cite{Lakshminarayanan2017}.

\subsection*{Data uncertainty}

One of the main sources of uncertainty -- and which is almost universally ignored -- is the uncertainty in the data, both in the predictors ($x$) and in the outcome to be predicted ($y$). The uncertainty can be in the training data used to build the model, in the new data to be predicted, or both. The main sources of data uncertainty are measurement error, misclassification error, binning, censoring and truncation, and missing values data. Each of these are discussed below.

Predictors are often experimental measurements and are therefore subject to measurement error, or they are samples from a larger population and are therefore subject to sampling error (e.g. only cells in the field of view are measured, not all cells in a well, and if a different subset of cells were selected, a differnt measured value would be obtained). Predictors may also be calculated quantities such as IC$_{50}$ values estimated from dose-response or concentration-response curves, and hence are uncertain. Furthermore, some predictors such as cLogP are the output of other (imperfect) prediction models and therefore are also uncertain. Finally, some predictors are not measured directly but are estimated from a standard curve, which introduces additional uncertainty because the curves may not be not perfectly calibrated. 

All these are examples of classic measurement error, defined as $x_{\mathrm{measured}} = x_{\mathrm{true}} + \mathrm{error}$, where the measured $x$ value is the true value corrupted by some error or noise. Errors can also be multiplicative, where $x_{\mathrm{measured}} = x_{\mathrm{true}} \times \mathrm{error}$. Another type of error is Berkson error, where samples or experimental units are assumed to have the same exposure but actually differ. For example, several wells in a microtitre plate are all given the same concentration of a compound, but in practice the true concentration may differ due to variations in the amount of compound dispensed or if wells closer to the edge of a plate have evaporated and thus have a higher effective concentration. Compound toxicity classifications can also introduce Berkson error. A compound may be classified as ``severely'' hepatotoxic, even though most people tolerate the compound well and only a few experience severe reactions. The class label is therefore defined by a few members of the class instead of the majority response. 

Berkson error often results from converting continuous values into bins or groups. For example, compounds are categorised as active versus inactive, despite having a range of activity values. Or, compounds are classified as having no, mild, or severe toxicity, even though compounds will have a range of toxicity levels \textit{within} each category. Binning can also lead to misclassification error, where a compound is placed into the incorrect category. This can occur if the measured assay value differed from the true value and fell on the opposite side of a threshold. Hence, binning is strongly discouraged \cite{Lazic2018a,Lazic2020}. Misclassification can also occur due to incorrect diagnoses, labelling errors, or data-entry errors. The standard response to these known and often large sources of data uncertainty is to ignore it, effectively assuming that $x_{\mathrm{measured}} = x_{\mathrm{true}}$.

It is well known that ignoring error in $x$ can bias parameter estimates, but for prediction models the parameters are usually not of interest. Even though noisy data can lead to biased parameter estimates, the model is still consistent for the prediction, meaning that as the sample size increases, the prediction will be correct, on average \cite{Gustafson2004,Carroll2006}. This likely explains why error in $x$ has received little attention in the predictive modelling and machine learning literature. However, if we're interested in the uncertainty in the prediction, then making good predictions on average is not good enough, we need to ensure that the prediction uncertainty is calibrated. 

Data are \textit{censored} when they are known only up to a boundary value, but not beyond, and therefore only partial information is available. For example, assays typically have upper and lower limits of detection (LoD) and uncertainty arises because the exact value is unknown, but it is typically treated as the ``true'' measured value.

Data are \textit{truncated} when values outside of a range are omitted, and the number of omitted values is unknown. For example, for objects to be segmented as a cell in a standard image analyses, they must have a minimum cell size. Smaller cells will not be included in the analysis, and both the estimated cell size and properties that are correlated with cell size can differ from their true value, thus introducing both uncertainty and bias.

Missing data is the final source of data uncertainty and can arise for many reasons. Imputation is a common approach to deal with missing data, where a plausible value is generated and substituted for the missing value. The imputed value is then taken as the true value, ignoring that it was generated and not measured. A simple way to account for uncertainty in an imputed value is to impute many values -- known as multiple imputation -- and the variation in the imputed values captures the uncertainty \cite{vanBuuren2012,Little2020}. Predictions are the made for each imputed value and the predictions combined.

\begin{figure}[ht]
  \centering
  \includegraphics[scale=0.44]{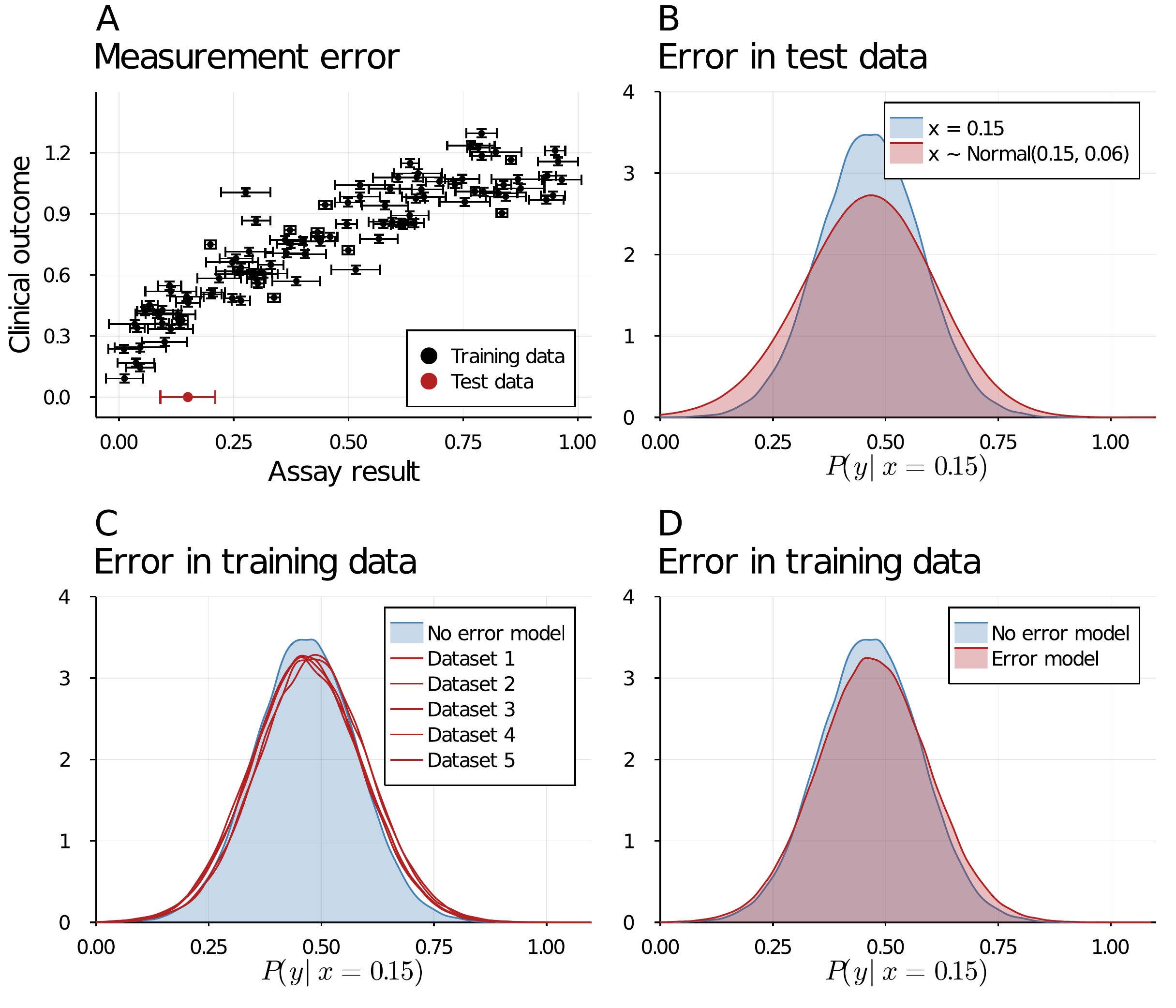}
  \caption{Data uncertainty. Error bars indicate 1 standard error of the estimated assay value and clinical outcome (A). The red point at $x=0.15$ is the new value to be predicted. Predictions are more uncertain when measurement error in the test data is included (assuming the training data is measured without error; B). Predictions for the test data when accounting for measurement error in the training data using multiple generated data (assuming no error in the test data; C). Averaged predictions from the generated data shows greater prediction uncertainty compared with ignoring measurement error in the training data (D).}
  \label{fig:x_err}
\end{figure}

To illustrate data uncertainty, Figure \ref{fig:x_err}A shows the same simulated data but with uncertainty in both the predictor and outcome variable. Here we assume that the error in $x$ differs for each sample because it depends on the precision of the measurement, whereas the uncertainty in $y$ is constant; for example, we know that the measurements are accurate to $\pm$ some fixed amount. Variable and fixed uncertainty are accounted for in the same way, and we include both for illustration purposes.

The uncertainty in $x$ and $y$ can be handled in two ways. The first is to directly model the errors in a Bayesian analysis \cite{Richardson1993,Gustafson2004,Muff2014,McElreath2016}. For example, if $x$ is measured with error, the true value can be inferred with $x_\mathrm{measured} \sim \mathrm{N}(x_\mathrm{true}, \sigma_\mathrm{error})$ where $\sigma_\mathrm{error}$ is the uncertainty in the measured value of $x$ -- usually the standard error. However, fitting such models may be difficult as each sample (row) adds as many parameters as there are variables with measurement error (columns).

A second simpler option is to generate multiple data sets, where the $x$ and $y$ values are drawn from a distribution \cite{Blackwell2015}. For example, suppose an $x$ variable for one compound has a measured value of 0.5 with a standard error of $\pm 0.06$. We can sample, say, 10 values from a normal distribution with a mean of 0.5 and a standard deviation of 0.06, thereby generating 10 new datasets where the observed value of $x$ is replaced with the generated value. This is a form of multiple imputation and Blackwell, Honaker, and King describe a more sophisticated method of generating new data by taking the correlations between variables into account \cite{Blackwell2015}. Each dataset is then analysed separately, and the predictions from each analysis are combined. Variations between the different datasets will lead to different parameter estimates, which in turn will lead to different predictions, and the ensemble of predictions captures the uncertainty in $x$. The models can be fit to the separate datasets in parallel, and so the computation time is the same as fitting the model to one dataset.

We use the second method to illustrate the effect of ignoring measurement error in two ways. First, we assume the training data is measured without error, but the test data is measured with error. Then we assume that the training data is measured with error but the test data is not. In both cases we compare the result to the standard approach of ignoring measurement error in both the training and test data. The test data is a single new compound with an assay value of $x=0.15 \pm 0.06$, shown as the red point in Figure \ref{fig:x_err}A (plotted at $y=0$, but the objective is to predict $y$). The 3-parameter exponential model is used in this example. The narrow blue distribution in Figure \ref{fig:x_err}B corresponds to the standard approach of ignoring uncertainty in both the training and test data, and the red distribution shows the greater uncertainty in the prediction for $y$ when the measurement error in $x$ is included. To obtain this distribution, 1000 samples were drawn from a normal distribution with a mean of 0.15 and standard deviation of 0.06. A prediction was made for each of these 1000 samples which reflects the uncertainty in $x$.

The blue distribution in Figure \ref{fig:x_err}C is again the standard analysis, and the five red lines show the slightly different predictions from each of the five datasets that account for measurement error in the training data (the test data was assumed to be error-free). Predictions from the five datasets are averaged and shown as the red distribution in Figure \ref{fig:x_err}D, which is slightly wider than the standard analysis from the blue distribution. The additional uncertainty appears negligible in this example, especially compared with uncertainty in the test data (Fig. \ref{fig:x_err}B). However, the variation between datasets is expected to increase with (1) greater uncertainty in the variables, (2) more variables with measurement error included in the model, and (3) more parameters in the model with the total sample size remaining fixed. The effect of measurement error in the training data can be assessed during model development and validation, and if the additional prediction uncertainty is negligible, then the final production model might ignore it.

\subsection*{Distribution function uncertainty}

Another source of uncertainty is the distribution function $G(\cdot)$, which represents our uncertainty in a predicted value of $y$.  Dozens of distributions are available but the list can be narrowed down based on background knowledge of the outcome. For example, if the outcome is binary such as absent/present, safe/toxic, or alive/dead, then a binomial distribution is appropriate; if the outcome is a count such as the number of seizures, then a Poisson or negative binomial distribution are two common options; if the data are positive values and skewed such as liver enzyme levels, then a log-normal or gamma distribution may be suitable; if the outcome is an ordered category such as none/mild/severe, then an ordered categorical distribution would be appropriate; if the outcome is continuous and unbounded with no outliers, then a Gaussian distribution may be suitable; and if there are outliers, a Student-t distribution might be appropriate. Many Bayesian textbooks have appendices that list the common distributions and their properties \cite{Gelman2004,Lesaffre2012,Lunn2013}.

Choosing between distributions is made easier because many distributions are special cases of other distributions. For example, both the Gaussian and Cauchy distributions are special cases of the Student-t distribution, the Poisson distribution is a special case of the negative binomial distribution, and the exponential distribution is a special case of a gamma distribution. Hence, we often don't need to choose between a set mutually exclusive options, but can select the more general distribution and allow the model to determine if one of the special cases is more appropriate. The more general distributions usually have only one additional parameter and therefore do not make the model much more complex. However, not all potentially suitable distributions are related (e.g. gamma and lognormal) and hence two or more models may need to be compared. The data for our running example was generated from a Gaussian distribution and which we have been using for all the models throughout. Hence using the more general Student-t distribution will inform us that the Gaussian is suitable, and so the results are not shown.

\begin{figure}[ht]
  \centering
  \includegraphics[scale=0.5]{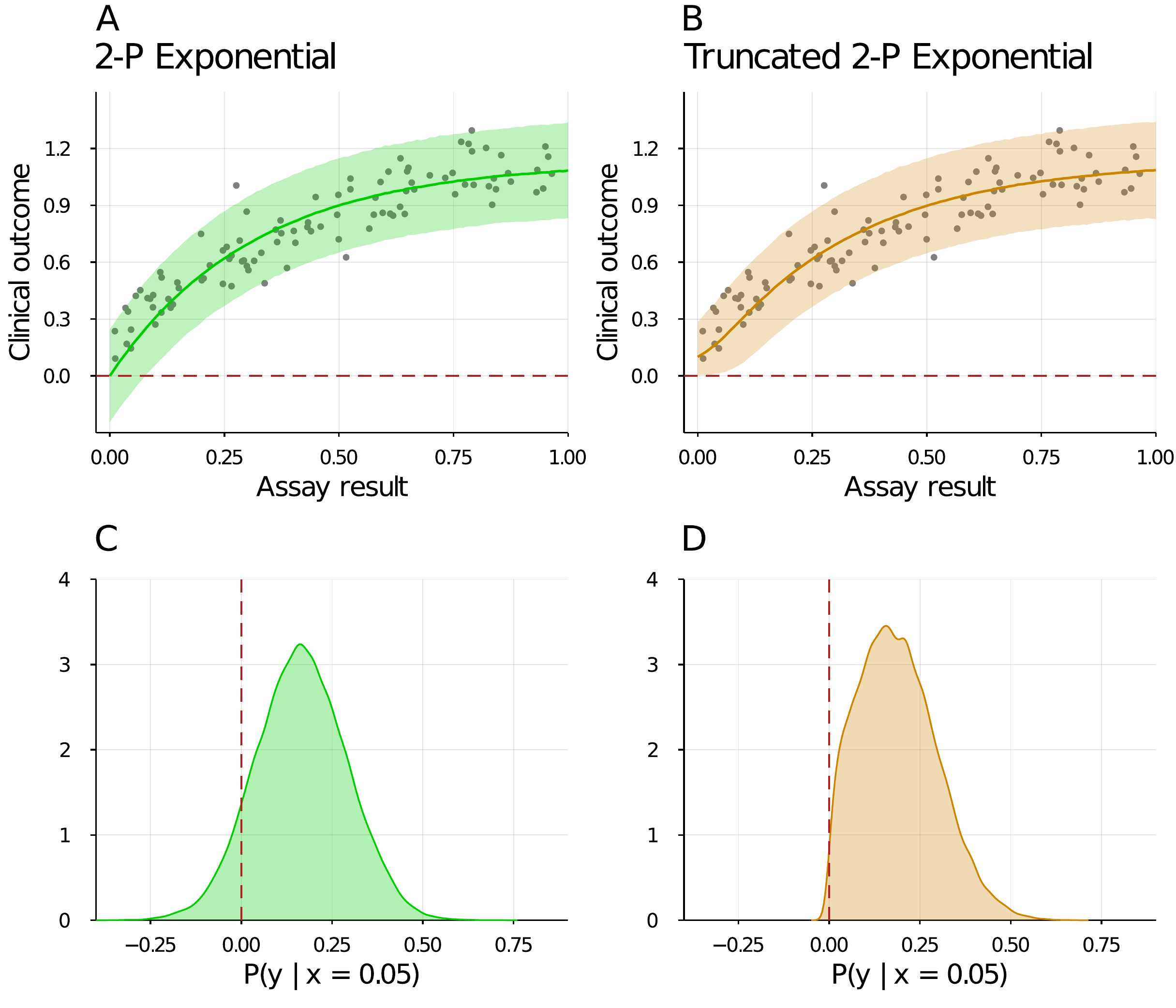}
  \caption{Truncated predictive distributions. Fits and 95\% PI for models  without (A and C) and with (B and D) a constraint that the predicted values must be positive. Prediction for a new compound given $x = 0.5$ without the constraint shows that 9.1\% of the predicted distribution is negative (C), while the truncated model redistributes the prediction to positive values (D). Red dashed lines are the data boundary.}
  \label{fig:trunc}
\end{figure}

A key consideration when selecting a distribution function is the bounds of the data. In our running example, the clinical outcome has a minimum value of zero, but is being modelled with a Gaussian distribution. Since a Gaussian distribution is defined for both positive and negative numbers, there is nothing to prevent negative predictions. A model is clearly inappropriate if it predicts impossible values. Fortunately, we can easily define truncated versions of standard distributions, and so we could specify a Gaussian distribution with a lower bound of zero. Figure \ref{fig:trunc} shows an example using the 2-parameter exponential model to predict the clinical outcome for an assay value of $x = 0.05$ both without (A) and with (B) truncation at $y=0$. Without truncation the model gives 9\% chance that $y$ will be less than zero (Fig. \ref{fig:trunc}C). With truncation, this probability gets redistributed to positive values (Fig. \ref{fig:trunc}D, the small proportion of the distribution below zero is a plotting artefact). Hence, if training or test data are near boundaries, using truncated versions of standard distributions is sensible. However, if the data are bounded but the values are far from the boundaries, then accounting for such boundaries may be unnecessary.

\subsection*{Link function uncertainty}

\begin{figure}[ht]
  \centering
  \includegraphics[scale=0.6]{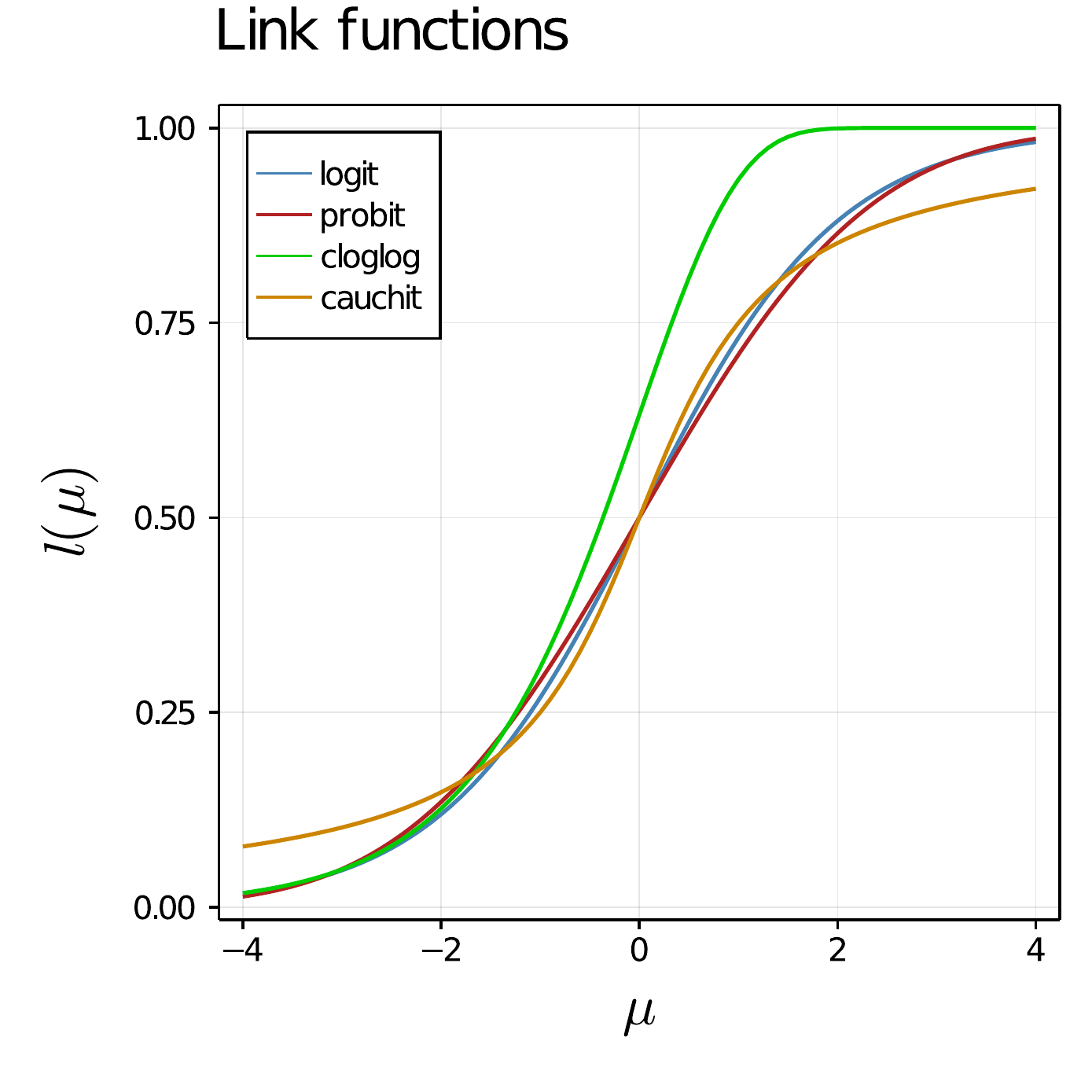}
  \caption{Link function uncertainty. Four functions that map the mean function $\mu = f(\cdot)$ from $-\infty$ to $\infty$ to values between 0 and 1. These link functions are required when the outcome is a probability and must be between 0 and 1. The predicted probabilities therefore differ depending on the link function.}
  \label{fig:link_funcs}
\end{figure}

Link functions are required when the predicted mean $\mu = f_\mu(\cdot)$ is bounded by an upper and/or lower limit. For example, when predicting the probability of an event, the predicted value must lie between 0 and 1. Values returned from the mean function are unconstrained and can lie well outside this range. Hence, a link function is used to transform the values to respect the bounds. The logit, probit, cauchit, and complementary log-log functions all take unconstrained numbers and compress them into the $0-1$ range (Fig. \ref{fig:link_funcs}). There is no ``correct'' link function and each provides a different mapping from the unconstrained input to the constrained output and hence gives a different prediction, especially for large values of the input. Link functions are also required for the variances, since variances cannot be negative values, and exponential or power links are often used.

Link functions are analogous to activation functions in neural networks, although they are used as nonlinear transformations between neurons and not necessarily to constrain values to allowable values. But the same issue arises in that many activation functions exist and different functions will lead to different predictions.

\subsection*{Variance function uncertainty}

\begin{figure}[ht]
  \centering
  \includegraphics[scale=0.6]{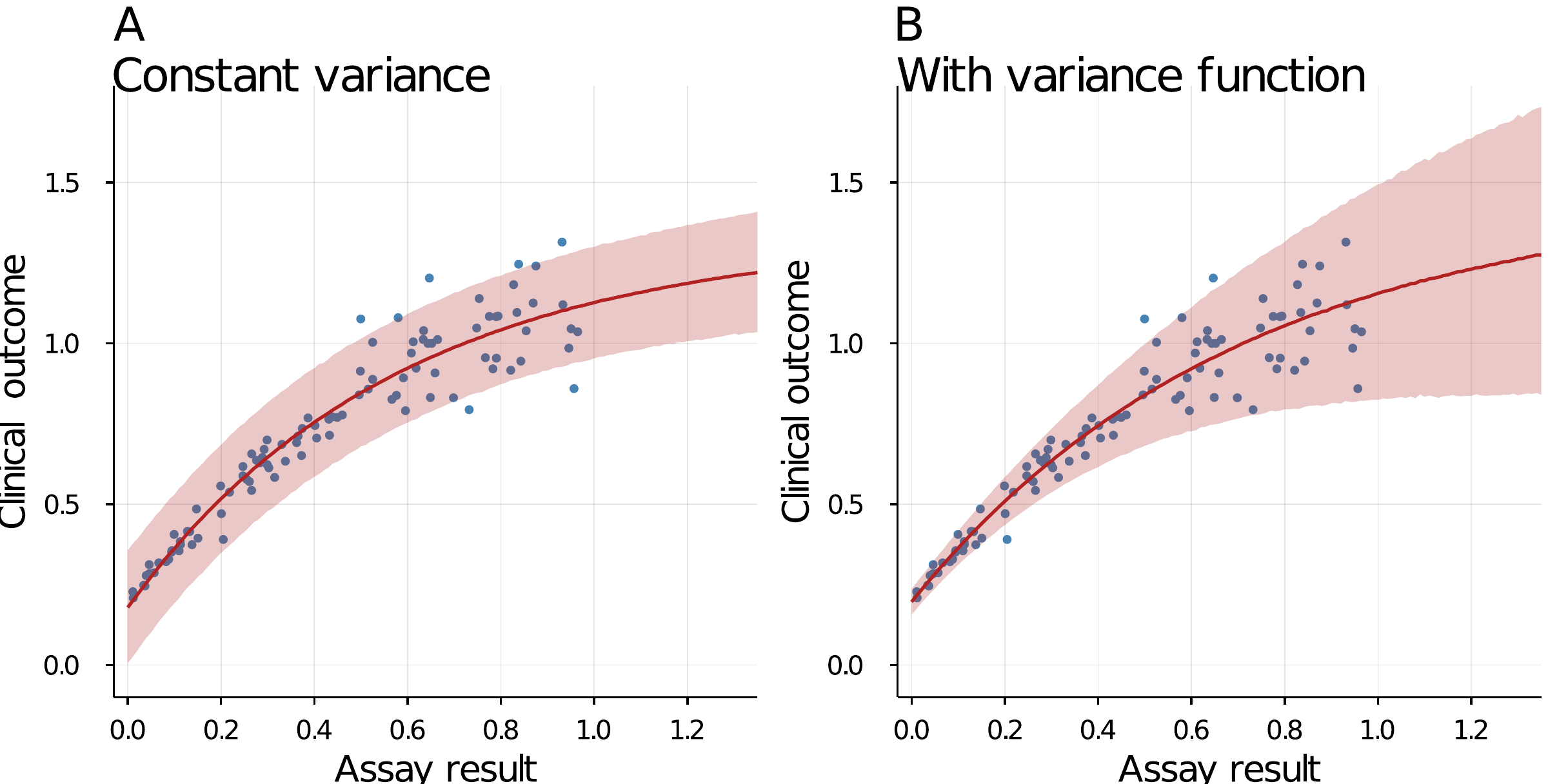}
  \caption{Variance function uncertainty. A constant variance implies that the uncertainty in a prediction is the same for all values $x$ (A). However, just like the mean prediction can change as function of $x$, so can the uncertainty in the prediction (B).}
  \label{fig:var_func}
\end{figure}

The variance function models the uncertainty in $y$ for given values of $x$. Another way to think of a variance function is that it models the spread of points around the mean prediction (Fig. \ref{fig:var_func}). The standard approach assumes that uncertainty in $y$ is constant (Fig. \ref{fig:var_func}A). However, when the variance is not constant, a model for $\sigma$ is required (Fig. \ref{fig:var_func}B). Just like modelling $\mu$ as a function of $x$, we now need to model $\sigma$ as a function of $x$. This function could be a simple function of one $x$ variable or a full neural network for all $x$ variables \cite{Nix1994}. The latter option involves creating a second neural network for the variance, but this doubles the complexity of the model and the training time. 

In Figure \ref{fig:var_func}B we do not use $x$ directly, but model $\sigma$ as a function of $\mu$ -- in other words, the uncertainty in $y$ is proportional to the predicted value of $y$. This allows for a simple mean function such as $f_{\sigma} = \sigma_0 + \sigma_1 \mu$, where $\sigma_0$ and $\sigma_1$ are parameters that control the relationship between $\sigma$ and $\mu$. Since variances must be positive values, a link function is needed to constrain $f_{\sigma}$ to be positive, and the softplus function $l_{\sigma} = \mathrm{log}(1 + \mathrm{exp}(f_{\sigma}(\cdot))$ is used here. The result is shown in Figure \ref{fig:var_func}B, where the 95\% shaded prediction region better matches the spread of the data compared with assuming a constant variance (Fig. \ref{fig:var_func}A).

Instead of using a variance function, another option is to transform the outcome variable (e.g. log, square-root, or inverse), so that the uncertainty in $y$ is constant. Alternatively, some distributions such as the Poisson and Bernoulli have a defined relationship between the variance the mean, which allows for non-constant variances, but they are only appropriate for certain type of data.

\subsection*{Sources of uncertainty combined}

Breaking down the sources of uncertainty into seven items enables us to think about them separately and assess their importance when developing a prediction model. A final model may include several sources and they can be easily combined. For example, suppose variance and link functions were not required but two mean functions and two distribution functions performed similarly and therefore four models with each combination of distribution and mean function are fit to the training data and the predictions averaged. If fully Bayesian models are used, parameter uncertainty is already account for. And if the test data are measured with error, we can use the approach in \ref{fig:x_err}B to draw multiple samples for each test sample and feed them all through the prediction models. The more sources of uncertainty accounted for the more complex the prediction model. Hence, sources of uncertainty that make little contribution to the overall prediction uncertainty can be ignored.

\section*{Uncertainty for classification tasks}

\begin{figure}[htb]
  \centering
  \includegraphics[scale=0.42]{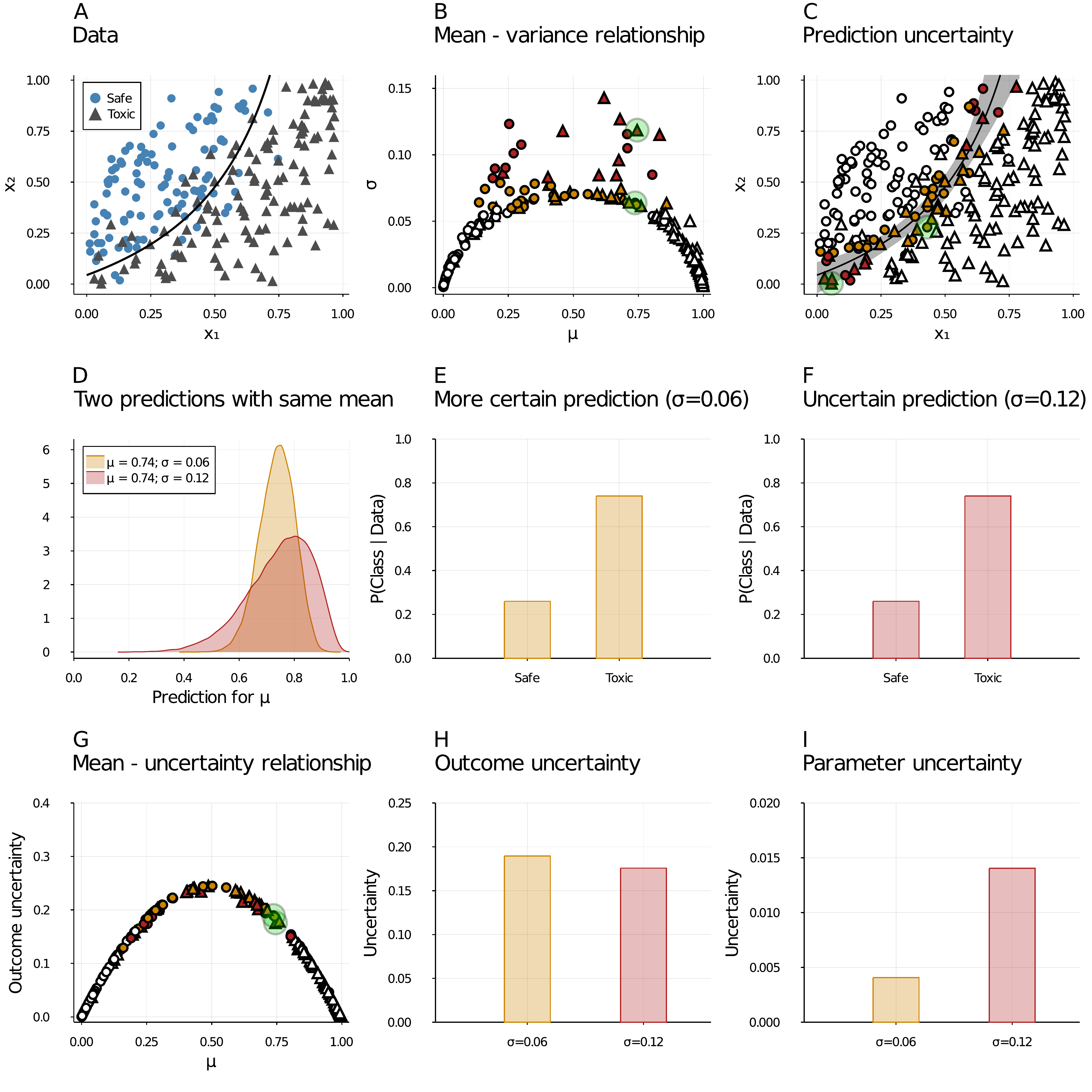}
  \caption{Uncertainty in classification. Simulated data with two features (A). A plot of the mean ($\mu$) and uncertainty ($\sigma$) shows the expected inverted-U relationship (B). The highest variance predictions (red points) are near the decision boundary and at the edge of the data, and high variance predictions (orange points) follow the boundary (C). Two compounds with the same mean but different uncertainties (D), have the same uncertainty in the final predicted value for $y$ (E,F). Outcome uncertainty is largely explained by $\mu$ (G) and is similar for the two compounds (H). Parameter uncertainty is nearly 3.5 time greater for the compound with the larger $\sigma$ (I).}
  \label{fig:class_uncert}
\end{figure}

The previous examples had a continuous outcome variable, but often outcomes are categorical such as toxic versus safe. Much of the previous discussion applies, but an important distinction is between uncertainty in a parameter ($\mu$) and uncertainty in a prediction for a new observable ($y$) \cite{Geisser1993,Briggs2016}. Greater parameter uncertainty leads to greater prediction uncertainty, but not for classification tasks, where the objective is to predict which of $K$ classes a sample belongs to. This point is illustrated in Figure \ref{fig:class_uncert}.

Figure \ref{fig:class_uncert}A plots data for a 2-group classification task with two predictors ($x_1, x_2$), and assume the grey triangles are the ``toxic'' class and the blue circles are the ``safe'' class. The black line is the optimal separating boundary. A logistic regression model is used to separate the classes and the prediction from the model will be a number between 0 and 1, where 1 corresponds toxic and zero corresponds to safe. This prediction is derived from the mean function and is passed through a link function to constrain the predictions to lie between 0 and 1. We'll call these predicted values $\mu$ and the uncertainty in the prediction $\sigma$. Figure \ref{fig:class_uncert}B plots $\mu$ versus $\sigma$ for each point in Figure \ref{fig:class_uncert}A, and the inverted-U relationship is a known feature of such models. But some samples are especially uncertain and are highlighted in red ($\sigma \geq 0.8$). Samples with intermediate uncertainty ($\sigma$ between 0.6 and 0.8) are highlighted in yellow. Figure \ref{fig:class_uncert}C shows that the uncertain samples are all close to the decision boundary, and that the most uncertain red points lie near the edge of the data where the location of decision boundary itself is uncertain. The shaded grey region in Figure \ref{fig:class_uncert}C represents the uncertainty in the decision boundary, and note how the uncertainty is wider at the ends compared with the middle.

\FloatBarrier

Figure \ref{fig:class_uncert}D plots the full distribution for two samples that have the same $\mu$, but one has twice the uncertainty. Recall that these distributions are for the parameter $\mu$ and not for the observable outcome $y$, which is either safe or toxic. To obtain a prediction for the observable we need a data generating distribution such as the Bernoulli distribution, giving us $y \sim \mathrm{Bernoulli}(\mu)$. Figure \ref{fig:class_uncert}E and F shows the predicted values of $y$ for these two compounds and note that they are identical, despite different values of $\sigma$. At first this may seem strange, but it is the desired behaviour. Consider tossing a coin four times and getting 3 heads, our prediction for the probability of heads is $3/4 = 0.75$ (and $1/4 = 0.25$ for tails). Similarly, if we toss a coin 1000 times and get 750 heads, our prediction is still 0.75, even though we are much more certain that the proportion of heads is 0.75 (i.e. $\sigma$ is much smaller). The width of the distributions in Figure \ref{fig:class_uncert}D (captured by $\sigma$) provides the \textit{weight of evidence} \cite{Keynes1921} which quantifies how much the prediction will change as more data are gathered or as the inputs change. For example, a single new observation with four coin tosses alters the probability to either $3/5 = 0.6$ or $4/5 = 0.8$, depending on whether a heads or tails was observed, whereas with 1000 tosses the probability is still 0.75, rounding to two decimal places.

Uncertainty in the predicted classes is often divided into aleatoric and epistemic uncertainty \cite{Kiureghian2009,Kendall2017}. Aleatoric uncertainty is supposedly due to ``inherent randomness'' whereas epistemic uncertainty is due to a lack of knowledge. Without starting a philosophical debate, we take the position that all uncertainty is due to a lack of knowledge \cite{Keynes1921,Briggs2016}. Nevertheless, we can decompose our uncertainty into two components, which we call the \textit{outcome uncertainty} and \textit{parameter uncertainty}, and which correspond to aleatoric and epistemic uncertainty, respectively. The first component is the distance of the point prediction to zero or one -- a more confident prediction would be close to these bounds, and the most uncertain prediction would be 0.5. This corresponds to outcome uncertainty. The second component is how confident we are in our point prediction. When a weatherperson states there is a 70\% chance of rain tomorrow, they do not mean exactly 70.00000\% but 70\% $\pm$ some amount. The uncertainty in the stated value corresponds to parameter uncertainty and is represented by the width of the distributions in Figure \ref{fig:class_uncert}D and the parameter $\sigma$. Using the approach of Kwon et al. we decompose the two sources of uncertainty for the compounds and plot outcome uncertainty versus the mean prediction ($\mu$) in Figure \ref{fig:class_uncert}G \cite{Kwon2020}. Note how $\mu$ largely explains the outcome uncertainty. The two compounds have similar outcome uncertainty since they have a similar value of $\mu$ (Fig. \ref{fig:class_uncert}H). However, the compound with the larger value of $\sigma$ has nearly 3.5 times greater parameter uncertainty (Fig. \ref{fig:class_uncert}H). Parameter uncertainty or the weight of evidence ($\sigma$), provides important information about the uncertainty of a prediction, which is critical for high-stakes decisions.

\section*{Discussion}

\subsection*{Generalisations and extensions}

The above examples used simple models but this framework can be generalised to more complex cases. For example, we had functions for the mean and variance, but any parameter in the distribution function can be modelled. For example, a Student-t distribution has a parameter called the degrees of freedom (df) which controls the heaviness of the tails. The df could be modelled as a function of $x$, just like $\mu$ or $\sigma$ \cite{Rigby2020}.

The above examples used a single distribution function, but flexibility can be increased by using mixtures of distributions. For example, outliers can be modelled with a mixture of Gaussian distributions: one to account for the regular observations and the second to account for the outliers. Metabolite, gene, and protein levels are non-negative and often positively skewed, and hence gamma or lognormal distributions may be appropriate. But these distributions are only defined for values \textit{greater} than zero, and there may be zeros in the data, which are often dealt with by adding a small value to all data points. A better option can be to model the data with a two-part model, one which accounts for the zeros and the other (e.g. gamma or lognormal) which accounts for the non-zero values. Such ``hurdle models'' provide this flexibility and also return a parameter that estimates the proportion of zeros, which may be scientifically interesting \cite{Cragg1971}. Taking this idea a step further, Dirichlet Process models allow us to specify as many distributions as needed to model the data. Instead of specifying a single distribution, we specify a prior over distributions, and learn them from the data (yes, we can specify a distribution over distributions! \cite{Mueller2015}).

The above examples also used a single mean function for each model, but it's possible to have a distribution of mean functions, which are called Gaussian Process models \cite{Rasmussen2006,Schulz2018,Gramacy2020}. These flexible models can fit complex relationships between $x$ and $y$. Surprisingly, they are not implemented via the mean function, but by generalising the variance function to make it a covariance function. Covariance functions are not discussed here but they are also useful for modelling hierarchical or nested data \cite{Johnstone2016,Pinheiro2000}, and for modelling dependencies in time or space. Neural networks \cite{Vehtari2000,Semenova2020,Hirschfeld2020} and Bayesian additive regression trees (BART) \cite{Chipman2010,Sparapani2016} are other options for flexible mean functions.


\subsection*{Further advantages of PPMs}
In addition to providing prediction uncertainty, PPMs have several other benefits. Hyperparameter values are typically selected by trying many options and choosing the combination that performs best. To avoid overfitting, crossvalidation or a similar approach divides the training data into smaller subsets, some of which are used for training and others to assess performance. But with small datasets, crossvalidation can give unstable models and a poor assessment of performance. Many Bayesian approaches can learn values of some hyperparameters using all the training data and have a built-in prevention of overfitting \cite{Tipping2001,Williams2020}. They also incorporate the uncertainty in the hyperparameters in the predictions. Models can still be compared using only the training data by estimating leave-one-out (LOO) crossvalidation performance, without the computational cost of actually retraining the model for each sample \cite{Vehtari2016,Vehtari2016a}. Vehtari and colleagues have also developed methods to assess when a LOO estimate is unreliable, and the model can be retrained only for these samples \cite{Vehtari2015}.

Another advantage is that background information such as adverse outcome pathways \cite{Burgoon2019}, constraints on parameters \cite{Lazic2018}, or monotonic relationships \cite{DePalma2017} can often be incorporated into the model, which can guide the model to better solutions.

Often several structurally similar compounds are available that have different binding affinities or potencies, but also with different results in the toxicity assays, and a decision must be taken to designate one compound in the series as the lead. PPMs can not only rank compounds but also obtain an uncertainty in the ranking, thus enabling decisions makes to conclude that one compound is reliably better than another \cite{Lazic2018,Semenova2020a}.

Finally, many popular machine learning methods have a PPM or Bayesian analogue, including regularised linear and generalised linear models (lasso, ridge regression) \cite{Park2008,Carvalho2009,Piironen2017,Rockova2018}, tree models (random forests, xgboost) \cite{Chipman1998,Chipman2010,Sparapani2016}, support vector machines \cite{Tipping2001,Sollich2002}, and neural networks \cite{Neal1996,Vehtari2000,Semenova2020}. Hence, it is often possible to convert your favourite model into one that provides prediction uncertainty.

\subsection*{Drawbacks and challenges}
The main drawback of Bayesian or other PMMs is that they require more work, possibly twice as much, since getting appropriately calibrated uncertainty is just as hard as getting accurate predictions. For example, 95\% prediction intervals should contain 95\% of the out-of-sample or test data values. \cite{Zhang2019}.

For fully Bayesian methods, the computational overhead may be high, making it difficult to iteratively fit, check, and update models during development (although computations are often much quicker when making predictions). Storage for parameter values may be a problem for large models since this equals the number of parameters times number of Markov chain Monte Carlo draws. These approaches may therefore be harder to scale to large datasets, but faster and scalable algorithms is an active area of research. Another solution to large data is to cleverly select a weighted subset of samples that is much smaller than the original but captures the essential features. This ``coreset'' approach enables standard PPM methods to be used on the smaller dataset with little loss of information \cite{Huggins2016,Campbell2018}.

Finally, not all sources of uncertainty can be captured. Many sources of uncertainty discussed above arise because many modelling options are available, and different choices lead to different predictions. All of the choices relate to the prediction model, but many decisions need to be made outside of the model. We refer to these extra-model choices as the workflow and they include experimental decisions such as the technology, cell-line, assay, antibodies, protocol, and so on. Also included are data processing pipelines where raw data are cleaned, transformed, categorised, coded, and normalised before they are entered into a prediction model. A single workflow is commonly used, with the untested assumption that variations in the workflow will lead to the same predictions and results. However, variations in workflows and analytic decisions do lead to variations results \cite{Shi2010,StantonGeddes2014,Steegen2016,Silberzahn2018,BotvinikNezer2020,Landy2020,HuntingtonKlein2021}.

\subsection*{Reporting uncertainty to help risk communication and decision making}
The ultimate aim of prediction models in drug discovery is to enable better decision making. Thus, not only should predictions be accurate with prediction uncertainty adequately represented, but the results should be easy to understand by decision makers. Fortunately, PPMs provide intuitive results for continuous (Fig. \ref{fig:trunc}D), binary (Fig. \ref{fig:class_uncert}D), categorical, and ordered categorical outcomes \cite{Williams2020,Semenova2020}, as well as for compound rankings \cite{Lazic2018,Semenova2020a}. We have found that safety pharmacologists and other project members can easily interpret the predictive distributions provided by PPMs and value the confidence in the predictions that these distributions provide \cite{Lazic2018,Williams2020}.

With recent advances in algorithms, hardware, and software, Bayesian or other PPMs are now feasible for most -- if not all -- machine learning problems encountered in drug discovery. Making PPMs the standard approach for critical ML problems will enable more informed and better decisions.


\bibliographystyle{model1a-num-names}

\end{document}